\documentclass[letterpaper]{article}
\usepackage{aaai}
\usepackage{times}
\usepackage{helvet}
\usepackage{courier}
\usepackage{amsmath,amssymb,amsfonts}
\usepackage{algorithmic}
\usepackage{graphicx}
\usepackage{multirow}
\usepackage{hyperref}

\begin{document}
%
\title {Equitable Allocation of Healthcare Resources with Fair Cox Models}
\author{ Kamrun Naher Keya,$^1$ Rashidul Islam,$^1$ Shimei Pan,$^1$ Ian Stockwell,$^2$ {and}  James Foulds$^1$\\ 
$^1$Department of Information Systems\\
$^2$The Hilltop Institute\\
University of Maryland, Baltimore County\\ 
\{kkeya1, islam.rashidul, shimei, jfoulds\}@umbc.edu, istockwell@hilltop.umbc.edu
}
\maketitle
\begin{abstract}
Healthcare programs such as Medicaid provide crucial services to vulnerable populations, but due to limited resources, many of the individuals who need these services the most languish on waiting lists. Survival models, e.g. the Cox proportional hazards model, can potentially improve this situation by predicting individuals' levels of need, which can then be used to prioritize the waiting lists. Providing care to those in need can prevent institutionalization for those individuals, which both improves quality of life and reduces overall costs. While the benefits of such an approach are clear, care must be taken to ensure that the prioritization process is fair or independent of demographic information-based harmful stereotypes. In this work, we develop multiple fairness definitions for survival models and corresponding fair Cox proportional hazards models to ensure equitable allocation of healthcare resources. We demonstrate the utility of our methods in terms of fairness and predictive accuracy on two publicly available survival datasets.     

\end{abstract}
\section{Introduction} 
Publicly funded healthcare programs such as Medicaid provide crucial services to vulnerable populations. Most states have subprograms  within their Medicaid programs meant to serve specific target populations. These programs are known as ``waivers,'' since each state must ask the federal government to waive some portions of the original Medicaid statute in order to better serve their population. With this expanded authority, states can include coverage for services that are not covered under traditional Medicaid programs (such as home and community-based long-term care), expand the financial eligibility requirements for participation, and limit the enrollment of each program for cost containment purposes. Many waivers are built to serve older adults or individuals with developmental/physical disabilities better by keeping them out of institutional settings (such as nursing homes). Participation in these programs with more services, relaxed financial eligibility, and limited enrollment becomes a necessarily scarce resource in need of allocation. The traditional method of allocating spots in these programs is a ``first in, first out'' model, where the next individual to enter the program is the one who has been waiting the longest.

Artificial intelligence (AI) can potentially improve this situation by predicting individuals' risk of institutionalization, which can then be used to prioritize the list of individuals who would like to participate in the program but for whom a spot on the waiver is not available (also known as the ``waitlist'').  
On October 1, 2019, the Maryland Department of Health deployed an AI system which performs a needs-based prioritization of the Medicaid waitlist as a function of predicted time to institutionalization, i.e. admission to a nursing home.
\footnote{\url{https://tinyurl.com/yy3odnmq}} 
While the benefits of such an approach are clear, care must be taken to \emph{ensure that the prioritization process is fair}. AI models can have impacts with lawful, moral or ethical consequences when utilized to predict outcomes in societal, governmental, and public sector applications.  
Structural and systemic processes, often unfair and/or biased against certain groups of people, impact individuals' lives and and hence their data~\cite{barocas2016big}, for example based on age, race, gender, nationality, class or sexual orientation. Since systemic bias is inherent in data, machine learning models must account for this to avoid creating discriminatory decisions. In recent years, the machine learning (ML) community has conducted a notable amount of research on algorithmic bias~\cite{dwork2012fairness,hardt2016equality,kusner2017counterfactual,foulds2020intersectional} which aims to learn non-discriminatory predictive models by enforcing constraints in the training phase~\cite{goel2018non,chouldechova2017fair,kilbertus2017avoiding,bolukbasi2016man,zhao2017men}. 

Like any data that involves individuals from different demographics, health data is subject to bias in various manners, and the expanding amount and types of data that are accessible today can make it difficult to distinguish where bias can emerge~\cite{ferryman2018fairness}. The goal of this work is therefore to develop AI techniques for attenuating harmful bias in the allocation of healthcare resources.

To predict individuals' risk of institutionalization, a natural approach is to use survival models. The Cox proportional hazards (CPH) \cite{cox1972regression} model is particularly appropriate in this research, as the multiplicative relationship between covariates and risk aids explainability. It is crucial to ensure fairness in the risk prediction task to obtain fair healthcare resource allocation with survival models such as CPH. Though the fairness community has proposed various fairness definitions~\cite{feldman2015certifying,dwork2012fairness,hardt2016equality,kearns2018preventing,foulds2020intersectional} to measure different aspects of societal or demographic biases in AI systems, to the best of our knowledge there are currently no fairness definitions specific to survival models.  
In this paper, we propose multiple fairness definitions for survival models and develop corresponding fair learning algorithms. The models' risk scores can then be used to fairly prioritize the Medicaid waitlist.  

The main contributions of this work include:
\begin{itemize}
    \item To the best of our knowledge, this is the first investigation on fairness for survival models to ensure equitable allocation of healthcare resources. 
    \item We extend three categories of fairness definitions to measures bias in the survival analysis problem.
    \item We develop corresponding fair learning algorithms for CPH models to ensure fair risk predictions.
    \item We perform extensive experiments validating our models with regards to both fairness and accuracy on two publicly available survival datasets.  
\end{itemize}
\section{Background and Related Work}
In this section, we define survival data and the Cox proportional hazards model, a popular method for survival analysis. In addition, we discuss related work on fairness in AI.
\subsection{Survival Data}
Survival data \cite{katzman2018deepsurv,lee2018deephit} contains three pieces of information for each individual: 1) observed covariates/features $x$, 2) actual time of the event $T$, and 3) event indicator $E$. If an event, e.g. death, has occurred, $T$ corresponds to the elapsed time between when the covariates were first collected and the time of the event occurring, and the event indicator is $E=1$. If an event is not observed, T corresponds to the elapsed time between the collection of the covariates and the last contact with the individual subject, e.g. end of the study, and $E$ becomes $0$. In this scenario, the individual subject is said to be \emph{right-censored}. In standard regression models missing data such as right-censored data may typically be discarded.  In survival analysis, however, right-censored data (e.g. data on individuals who survive to the end of the study) is important and cannot simply be ignored without introducing substantial bias. Right-censored data therefore requires special consideration in this context.

\subsection{Cox Proportional Hazards Model (CPH)}
The Cox proportional hazards (CPH) model \cite{cox1972regression} is the most widely used model for survival analysis \cite{lee2018deephit}. It is  a semiparametric model often used in clinical (and many other) settings for modeling and predicting the time until a particular event occurs, e.g. death of a patient.  

Let $S(t)$ be the probability that the event does not occur before time $t$ (the \emph{survival function}). The key concept to define these models is the \emph{hazard function}, defined to be the instantaneous rate that the event, e.g. death or institutionalization, occurs at time $t$, here supposing that time is continuous. The hazard function is defined as
\begin{align}
\small
    h(t) \triangleq \lim_{\Delta t \rightarrow 0^+}\frac{Pr(t \leq T < t + \Delta t | T \geq t)}{\Delta t} \mbox{ .}
\end{align}
The CPH model specifies the hazard function via
\begin{align}
\small
    h(t) = h_0(t) \exp(\beta^\intercal \mathbf{x}) \mbox{ ,}
\end{align}
where $h_0$, called the \emph{baseline hazard}, is the hazard value regardless of features $\mathbf{x}$, and $\beta$ is a parameter vector.  The survival function is determined from the hazard function via
\begin{align}
\small
    S(x) = \exp(-H(t)), H(t) = \int_0^t h(u) du \mbox{ .}
\end{align}

To perform Cox regression, the parameter $\beta$ can be learned by optimizing the Cox partial likelihood \cite{faraggi1995neural,katzman2018deepsurv}. The partial likelihood is the product of the probability at each event time $T_i$ that the event $E_i$ has occurred to individual $i$, given the set of individuals still at risk at time $T_i$ and can be calculated as
\begin{equation}
    L_c(\beta) = \prod_{i:E_i = 1}\frac{exp(\beta^\intercal \mathbf{x}_{i})}{\sum_{j\in \Re (T_{i})}\exp(\beta^\intercal \mathbf{x}_{j})}\mbox{ ,} \label{eqn:likelihood}
\end{equation}
where the product is defined over the set of patients with an observable event $E_i = 1$ and the risk set $\Re(t)=\{i:T_i\geq t\}$ is the set of patients still at risk of failure at time $t$. 

The CPH model assumes that an individual's risk of an event occurring is a linear combination of the patient's covariates, referred to as the linear proportional hazards condition.  
Since this assumption may be too simplistic in many applications such as personalized treatment recommendations \cite{katzman2018deepsurv},
recently deep neural networks \cite{faraggi1995neural,katzman2018deepsurv,lee2018deephit,huang2019salmon} have been applied to CPH models to solve the problem of nonlinear survival analysis. 
\subsection{Fairness in AI}
The increasing impact of artificial intelligence (AI) and machine learning technologies on many facets of life, from commonplace movie recommendations to consequential criminal justice sentencing decisions, has prompted concerns that these systems may behave in an unfair or discriminatory manner \cite{barocas2016big,executive2016big,noble2018algorithms}. A number of studies have subsequently demonstrated that bias and fairness issues in AI are both harmful and pervasive \cite{angwin2016machine,buolamwini2018gender,bolukbasi2016man}. The AI community has responded by developing a broad array of mathematical formulations of fairness and learning algorithms which aim to satisfy them \cite{dwork2012fairness,hardt2016equality,berk2017convex,zhao2017men}.  
An overview of fair AI is given by \cite{berk2017fairness}.

While a number of fairness definitions have been proposed in the literature, at the highest level there are three broad categories of fairness measures.   \textbf{Individual fairness} definitions aim to ensure that \emph{similar individuals obtain similar outcomes} under the algorithm in question.   \textbf{Group fairness} definitions aim to preserve fairness at the level of groups of individuals, e.g. women, the elderly, or African Americans.  Finally, \textbf{intersectional fairness} definitions are those for which fairness is to be ensured for a specified set of subgroups defined by the protected attributes \cite{kearns2018preventing,hebert-johnson2018multicalibration,foulds2020intersectional}. These fairness definitions can be slightly modified to form a fairness penalty that can be added as  a constraint or a regularization term to the existing optimization objective to enforce fairness in the algorithm \cite{calders2010three,zafar2015fairness,zafar2017fairness,foulds2020intersectional}.

\cite{angwin2016machine} applied Cox models to the problem of criminal recidivism prediction, in order to detect racial bias in the offender's risk score prediction. However, there is no prior work that introduces and enforces formal fairness definitions for the fair survival analysis problem. 

\section{Methods}
In this section, we describe our methodology to ensure fair risk predictions with survival models. First, we extend the three broad categories of fairness measures to precise application of fairness in survival analysis problems. We then develop a simple learning algorithm for fair survival models. 

\subsection{Fairness Definitions for Survival Models}
Fairness in healthcare is a multi-stakeholder issue, and so we cannot simply settle it with a single solution.  We instead provide stakeholders with three different proposed implementations of fairness for survival models.

\textbf{Individual fairness:} 
Individual fairness \cite{dwork2012fairness} aims to ensure that a system or model produces similar outcomes to similar individuals. 
In the context of survival models, we define individual fairness ($F_i$) inspired by \cite{dwork2012fairness}  as follows: 
\begin{align}
\small
    F_i = \sum_{i=1}^{N^{(test)}}\sum_{j=i+1}^{N^{(test)}}  \max(0, |\bar{h}_\beta(\mathbf{x}_i) - \bar{h}_\beta(\mathbf{x}_j)| - D(\mathbf{x}_i,\mathbf{x}_j)) \mbox{ ,} \label{eqn:coxIndividual}
\end{align}
where $\bar{h}_\beta(\mathbf{x}) = \exp(\beta^\intercal \mathbf{x})$, the hazard function where the base hazard $h_0(t)$, which is not individual-specific, is dropped, and $D(x_i,x_j)$ is a distance metric (e.g. Euclidean distance) between $x_i$ and $x_j$ encoding fair similarity, on the same scale as $|\bar{h}_\beta(\mathbf{x}_i) - \bar{h}_\beta(\mathbf{x}_j)|$.  Note that this penalizes differences in predicted hazard scores that exceed the distance between the data points. Here, we can make use of knowledge of the individuals who are to be in the test set such as the individuals on the waitlist for care (a \emph{transductive} approach to fairness).

\textbf{Group fairness:} In group fairness definitions, e.g. demographic parity \cite{dwork2012fairness}, a system is fair if outcomes are distributed fairly across different demographic groups, e.g. different genders or races. We define the group fairness ($F_g$) measures for survival models as
\begin{align}
\small 
    F_{g} = \max_{a \in A} |\bar{h}_\beta(a) - E[\bar{h}_\beta(\mathbf{x})]| \mbox{ , }\\ \bar{h}_\beta(a) \triangleq \int_{\mathbf{x}} \exp(\beta^\intercal \mathbf{x}) p(\mathbf{x}|a) \mbox{ ,} \label{eqn:coxGroup}
\end{align}
the worst-case deviation of the \emph{per-group expected hazard function} $\bar{h}_\beta(a)$ from the population average hazard where $A$ is the set of values in the protected attribute.
We estimate the above integral via an average over the empirical data.

\textbf{Intersectional fairness:} Intersectional fairness \cite{kearns2018preventing,hebert-johnson2018multicalibration,foulds2020intersectional} definitions consider subgroups of protected groups, usually defined to be their intersecting subgroups.  This can be used to enforce fairness metrics that encode the principle of intersectionality \cite{crenshaw1989demarginalizing}, namely that individuals at the intersections of protected groups, e.g. along lines of race and gender, are vulnerable to additional harms and should be protected \cite{foulds2020intersectional}.  In this case, $A = S_1 \times S_2 \times \ldots S_K$ is a space of multi-dimensional protected attributes.  Building on our earlier work on the \emph{differential fairness} metric \cite{foulds2020intersectional}, intersectional fairness ($F_\epsilon$) for survival models can be extended as 
\begin{align}
\small
    F_{\epsilon} = \max_{s_i \in A, s_j \in A} |\log \bar{h}_\beta(s_i) - \log \bar{h}_\beta(s_j)| \mbox{ ,} \label{eqn:coxIntersectional}
\end{align}
a worst case of log-ratios of ``per-group hazard functions'' over pairs of intersectional subgroups $s_i$, $s_j$ (e.g. men over 70, women between 20 - 30). In this formulation, fairness for intersectional subgroups provably guarantees fairness for the higher-level groups \cite{foulds2020intersectional}.

\subsection{Fair Cox Proportional Hazards Models (FCPH)}
We develop simple and practical Fair Cox Proportional Hazards (FCPH) models which balance fairness and accuracy. FCPH models allow us to fair prediction of the time until a particular event occurs, for example, institutionalization into a nursing home. Therefore, the FCPH models' risk scores can be used to prioritize the ``waitlist" of patients for fair allocation of healthcare resources, independent of the patient's demographics, e.g. \emph{age}, \emph{gender}, \emph{race} etc. 

The linear Cox model estimates the hazard function $\hat{h}(t)$ parameterized by the weight vector $\beta$. Following \cite{faraggi1995neural,katzman2018deepsurv}, the loss function to learn $\beta$ can be formulated with the negative log partial likelihood of Equation~\ref{eqn:likelihood}:
\begin{equation}
    -L_{\mathbf{X}}(\beta) = -\sum_{i:E_{i}=1}(\beta^\intercal \mathbf{x}_{i}-log\sum_{j\in \Re (T_{i})}\exp(\beta^\intercal \mathbf{x}_{j})) \mbox{ .} \label{eqn:loglike_loss}
\end{equation}

Our FCPH models are developed upon a general framework for solving fairness in linear Cox models using a penalized maximum likelihood estimation approach.  The general learning objective function for our FCPH models is
\begin{align}
\small
    -L_{\mathbf{X}}(\beta) + \lambda R(\beta) \mbox{ , }\label{eqn:objective} 
\end{align}
where $L_{\mathbf{X}}(\beta)$ is the log-likelihood for the linear Cox model, $R_{\mathbf{X}}(\beta)$ is a fairness penality, which also doubles as a regularizer, and $\lambda>0$ is a trade-off parameter which strikes a balance between predictive accuracy and fairness. We set $R_{\mathbf{X}}(\beta)$ to $F_i$, $F_g$, and $F_{\epsilon}$ fairness measures to learn \emph{individual}, \emph{group}, and \emph{intersectional} \emph{FCPH} models, respectively. We optimize the objective function in Equation~\ref{eqn:objective} via gradient descent algorithm \cite{ruder2016overview} to learn models' parameter vector $\beta$. This approach is applicable to training deep Cox models as well, which we will study in future work. 

\section{Experiments} 
In this section, we validate and compare our fair Cox (FCPH) models with the typical Cox (CPH) model on the survival datasets for fair risk predictions which can be used to perform sensitive tasks such as prioritizing the waiting list for institutionalization in the healthcare programs.

\subsection{Datasets}
Data for the allocation of healthcare resources, e.g. prioritizing the wait list of patients for institutionalization, is not publicly available. Therefore, we validate our models on representative proxy datasets. We performed all experiments on two publicly available survival datasets:
\begin{itemize}
    \item \textbf{\emph{COMPAS} Data:}  The \emph{COMPAS} dataset regarding a system that is used to predict criminal recidivism, and which has been criticized as potentially biased ~\cite{angwin2016machine}. ~\citeauthor{angwin2016machine} applied a Cox model to test the performance of the \emph{COMPAS} system on offender's risk prediction and demonstrated that the system overpredicts African-American defendant's future recidivism. The company that developed COMPAS system and markets it to Law Enforcement, also used a Cox model in their validation study \cite{brennan2009evaluating}.
    Although the COMPAS system is used for bail and sentencing, it could potentially be used to allocate social work resources.  
    Therefore, it is a useful example dataset in our study to show the effectiveness of our methods to fair risk prediction. The \emph{COMPAS} dataset consists of $10,314$ offenders and $6$ features including demographic attributes, while the task is to predict risk scores of a convicted criminal to reoffend. A total of $26.75\%$ of subjects reoffended during the survey for data collection with a median event time of $173$ days. We used binary-coded \emph{race} (white, and African-American) and \emph{gender} (men, and women) as protected attributes in our study. 
    \item \textbf{\emph{FLC} Data:} This dataset is taken from a study that investigated to which extent the serum immunoglobulin free light chain (FLC) assay can be used predict overall survival \cite{dispenzieri2012use}. ~\citeauthor{dispenzieri2012use} assayed FLC levels on the patients with permission from a previous study for the prevalence of monoclonal gammopathy of undetermined significance \cite{kyle2006prevalence} and found that elevated FLC levels were indeed associated with higher death rates. The \emph{FLC} Dataset consists of $7,874$ patients with $6$ features such as age, gender, serum creatinine, FLC group for the patients, kappa and lambda portion for serum free light chain, while the task is to predict the risk score for death. A total of $27.55\%$ of patients died during the survey with a median death time of $2,165$ days. We used binary-coded \emph{age} (age$\leq65$, and age$>65$) and \emph{gender} (men, and women) as protected attributes in our fairness analysis.   
\end{itemize}
\begin{figure*}[t]
		\centerline{\includegraphics[width=0.99\textwidth]{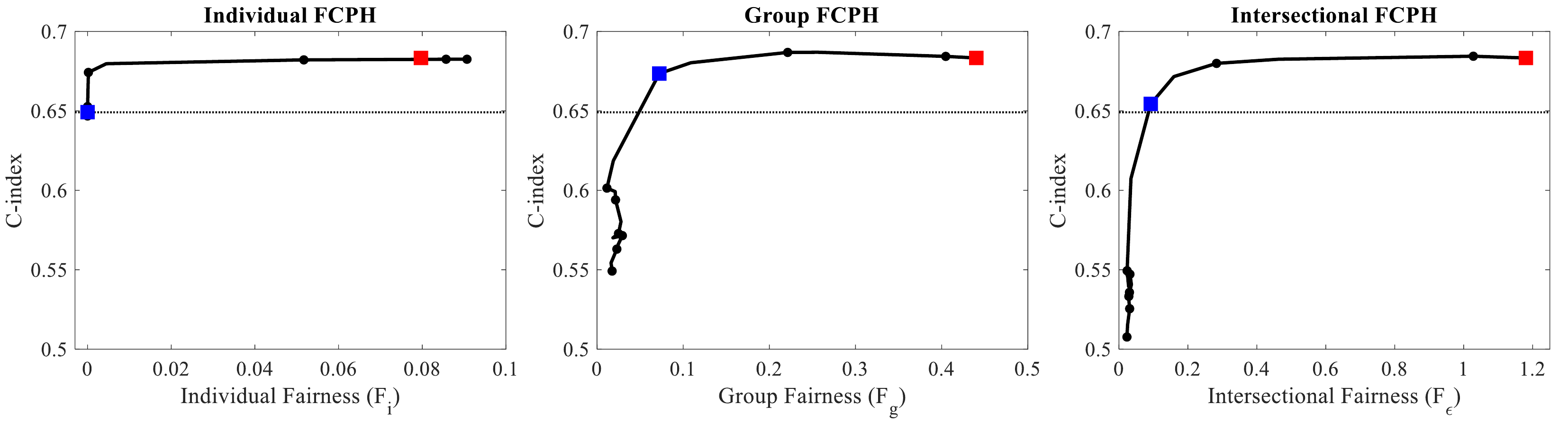}}
		\caption{\small Fairness and accuracy trade-off plots for the development set of the \emph{COMPAS} dataset. Shows the impact of tuning parameter $\lambda$ on the FCPH models' C-index and corresponding fairness measures. \emph{Black} circles correspond to different $\lambda$ values (larger to smaller from left to right), while \emph{blue} square indicates the selected FCPH model for a specific $\lambda$ value. \emph{Red} square: typical CPH model without fairness penalty. \emph{Dotted} line represents  $5\%$ degraded \emph{C-index} from the typical CPH model. C-index: higher is better.  Fairness measures: lower is better.}
		\label{fig:compas_lambda}
\end{figure*}
\begin{figure*}[t]
		\centerline{\includegraphics[width=0.99\textwidth]{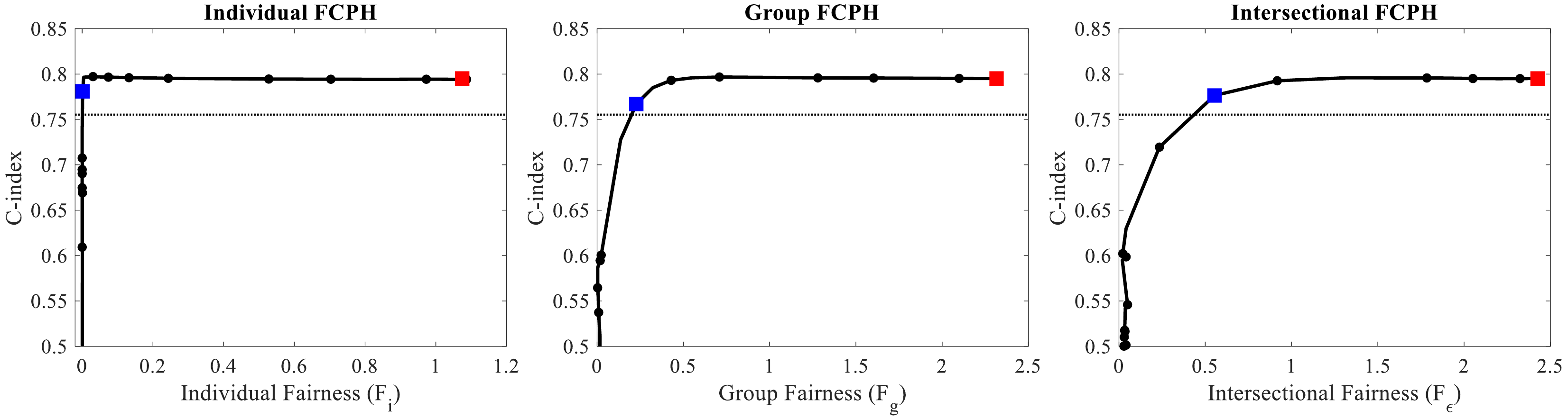}}
		\caption{\small Fairness and accuracy trade-off plots for the development set of the \emph{FLC} dataset. Shows the impact of tuning parameter $\lambda$ on the FCPH models' C-index and corresponding fairness measures. \emph{Black} circles correspond to different $\lambda$ values (larger to smaller from left to right), while \emph{blue} square indicates the selected FCPH model for a specific $\lambda$ value. \emph{Red} square: typical CPH model without fairness penalty. \emph{Dotted} line represents  $5\%$ degraded \emph{C-index} from the typical CPH model. C-index: higher is better.  Fairness measures: lower is better.}
		\label{fig:flc_lambda}
\end{figure*}
\begin{table*}[t]
\centering
\begin{tabular}{llcccc}
\hline
\multicolumn{2}{c}{Models}  & \multicolumn{1}{l}{Typical CPH} & \multicolumn{1}{l}{Individual FCPH} & \multicolumn{1}{l}{Group FCPH} & \multicolumn{1}{l}{Intersectional FCPH} \\ \hline
Tuning                       & $\lambda$               & X                           & 25                                  & 1                                     & 1                                       \\ \hline
\multirow{4}{*}{Performance} & C-index       & \textbf{0.6648}             & 0.6341                              & 0.6577                                & 0.6445                                   \\
                             & Brier Score             & 0.1877                      & 0.1823                              & 0.1808                                & \textbf{0.1787}                         \\
                             & Time-dependent AUC      & 0.6872                      & 0.6584                              & \textbf{0.6890}                       & 0.6745                                \\
                             & Log Partial Likelihood  & \textbf{-7.0954}            & -7.1664                             & -7.1167                               & -7.1542                               \\  \hline
\multirow{4}{*}{Fairness}    & $F_i$-Fairness     & 0.0968                      & \textbf{0}                          & 0.0090                                & 0.0004                                \\
                             & $F_g$-Fairness          & 0.4198                      & 0.1999                              & 0.0765                                & \textbf{0.0418}                      \\
                             & $F_{\epsilon}$-Fairness & 1.0821                      & 0.7291                              & 0.2421                                & \textbf{0.1067}                         \\ \hline
\end{tabular}
\textbf{\caption{\small Comparison of FCPH models with typical CPH model on the \emph{COMPAS} dataset. \emph{Higher is better for accuracy-based performance measures}; \emph{lower is better for fairness measures}. FCPH models outperform typical CPH model in terms of all fairness measures.}
\label{fig:compas}}
\end{table*}
\begin{table*}[t]
\centering
\begin{tabular}{llcccc}
\hline
\multicolumn{2}{c}{Models}  & \multicolumn{1}{l}{Typical CPH} & \multicolumn{1}{l}{Individual FCPH} & \multicolumn{1}{l}{Group FCPH} & \multicolumn{1}{l}{Intersectional FCPH} \\ \hline
Tuning                       & $\lambda$               & X                           & 5                                   & 0.7                                   & 0.4                                       \\ \hline
\multirow{4}{*}{Performance} & C-index       & \textbf{0.8030}             & 0.7898                              & 0.7768                                & 0.7885                                  \\
                             & Brier Score             & 0.2244                      & \textbf{0.1911}                     & 0.1951                                & 0.1976                         \\
                             & Time-dependent AUC      & 0.8015                      & \textbf{0.8173}                     & 0.8147                                & 0.8155                                \\
                             & Log Partial Likelihood  & \textbf{-6.3737}            & -6.9207                             & -6.7963                               & -6.7299                             \\ \hline
\multirow{4}{*}{Fairness}    & $F_i$-Fairness     & 1.4073                      & \textbf{0.0009}                     & 0.0243                                & 0.0343                                \\
                             & $F_g$-Fairness          & 3.0027                      & \textbf{0.1466}                     & 0.2879                                & 0.3959                     \\
                             & $F_{\epsilon}$-Fairness & 2.8334                      & \textbf{0.3761}                     & 0.7468                                & 0.6610                         \\ \hline
\end{tabular}
\textbf{\caption{\small Comparison of FCPH models with typical CPH model on the \emph{FLC} dataset. \emph{Higher is better for accuracy-based performance measures}; \emph{lower is better for fairness measures}. FCPH models outperform typical CPH model in terms of all fairness measures.}
\label{fig:flc}}
\end{table*}
\begin{table*}[t]
\centering
\begin{tabular}{lcccccc}
\hline
\multirow{2}{*}{Models} & \multicolumn{3}{c}{Train Set}  & \multicolumn{3}{c}{Test Set}   \\
                        & C-index & Brier Score & AUC    & C-index & Brier Score & AUC    \\ \hline
Typical CPH             & \textbf{0.6859}  & \textbf{0.1530}      & \textbf{0.7178} & \textbf{0.6648}  & 0.1877               & 0.6872 \\
Individual FCPH         & 0.6609           & 0.1646               & 0.6962          & 0.6341           & 0.1823               & 0.6584 \\
Group FCPH              & 0.6654           & 0.1610               & 0.7031          & 0.6577           & 0.1808               & \textbf{0.6890} \\
Intersectional FCPH     & 0.6593           & 0.1648               & 0.6964          & 0.6445           & \textbf{0.1787}      & 0.6745 \\ \hline
\end{tabular}
\textbf{\caption{\small \small Comparison of the accuracy-based predictive performances for typical CPH and FCPH models on the train and test set of the \emph{COMPAS} dataset. FCPH models reduce overfitting.}
\label{fig:compas_train}}
\end{table*}
\begin{table*}[t]
\centering
\begin{tabular}{lcccccc}
\hline
\multirow{2}{*}{Models} & \multicolumn{3}{c}{Train Set}  & \multicolumn{3}{c}{Test Set}   \\
                        & C-index & Brier Score & AUC    & C-index & Brier Score & AUC    \\ \hline
Typical CPH             & \textbf{0.7944}  & \textbf{0.1298}      & \textbf{0.8149} & \textbf{0.8030}  & 0.2244               & 0.8015 \\
Individual FCPH         & 0.7740           & 0.1719               & 0.8089          & 0.7898           & \textbf{0.1911}               & \textbf{0.8173} \\
Group FCPH              & 0.7581           & 0.1613               & 0.7961          & 0.7768           & 0.1951               & 0.8147 \\
Intersectional FCPH     & 0.7680           & 0.1555               & 0.8012          & 0.7885           & 0.1976      & 0.8155 \\ \hline
\end{tabular}
\textbf{\caption{\small \small Comparison of the accuracy-based predictive performances for typical CPH and FCPH models on the train and test set of the \emph{FLC} dataset. FCPH models reduce overfitting.}
\label{fig:flc_train}}
\end{table*}
\subsection{Experimental Settings}
All the models were trained via the adaptive gradient descent optimization (Adam) algorithm \cite{kingma2014adam} with learning rate $0.01$ using PyTorch. There is no requirement of protected attributes to learn the \emph{Individual FCPH} model since the individual fairness depends on each individual subject rather than any group/subgroup of peoples. We considered \emph{race} for \emph{COMPAS}, and $\emph{age}$ for \emph{FLC} datasets as protected attributes in the \emph{Group FCPH} model, while we considered all the pre-selected protected attributes (\emph{race}, \emph{gender} for \emph{COMPAS}, and \emph{age}, \emph{gender} for \emph{FLC} datasets) in the \emph{Intersectional FCPH} model.

We held out $20\%$ of each dataset as the test set, using the remainder for training. We further held out $20\%$ from each training dataset as the development set for each dataset. Since it is challenging to estimate group and intersectional fairness reliably on mini-batches due to data sparsity \cite{foulds2020bayesian}, we trained all the models, except the \emph{Individual FCPH} model, in a batch setting for $500$ iterations. It becomes very expensive to measure individual fairness on the whole training set in each iteration when training \emph{Individual FCPH} models in a batch setting. Furthermore, we found that data sparsity is not a serious issue for individual fairness measures, unlike group and intersectional fairness. Therefore, to address the bottleneck we trained the \emph{Individual FCPH} models in a mini-batch setting for $50$ epochs with a mini-batch size of $128$. 

\subsection{Evaluation Protocols}
In addition to the fairness measures we proposed (Equations \ref{eqn:coxIndividual}, \ref{eqn:coxGroup}, and \ref{eqn:coxIntersectional}), in our evaluation we also included traditional accuracy measures 
for the predictive performance of the survival models: \emph{Concordance Index (C-index)}, \emph{Brier Score}, \emph{Time-dependent AUC}, and \emph{Log Partial Likelihood}.   
The \emph{C-index} ~\cite{Raykar07,Uno2011,Brentnall2018} is a rank order statistic for predictions against true outcomes, thus highly relevant for waitlists. It is a generalization of the Area Under the ROC Curve (AUC) for continuous response and censored data that reflects a measure of how well a model predicts the ordering of individual's event time. The \emph{C-index} is 
based on the assumption that patients who lived longer should have been assigned a lower risk than patients who lived less long~\cite{harrell1982evaluating}. A high \emph{C-index} means that there is a high likelihood that for two random samples, the order of their predicted response matches the order of their observed response.

The \emph{Brier Score}~\cite{Graf1999} measures the accuracy of probabilistic predictions. Given a set of $N$ predictions, the empirical Brier Score measures the weighted mean squared difference between the predicted probability assigned to possible outcomes for sample $i$ and the actual outcome. The weight for sample $i$ can be estimated by considering the Kaplan-Meier estimator~\cite{Kaplan1958} of the censoring distribution on the dataset. 

The \emph{Time-dependent AUC}~\cite{chambless2006estimation} is a function of time that extends the ROC curve to continuous outcomes, in particular survival time, assuming a subject’s event status is typically not fixed and changes over time, e.g. patients who are disease-free earlier may develop the disease later due to longer study follow-up. It reflects the area under the cumulative/dynamic ROC at time $t$ to determine how well a model can distinguish subjects that experienced an event prior to or at time $t$ (cumulative cases) from subjects that experienced an event after this time point (dynamic controls). 

Finally, we used the \emph{Log Partial Likelihood} from Equation~\ref{eqn:loglike_loss} (dropping the minus sign) as a performance measure.  The effect of the covariates can be estimated using the \emph{Log Partial Likelihood} without the need to model the change of the hazard over time and it measures the goodness of fit of models to a sample of data for learned parameter $\beta$.

\subsection{Trade-off Between Fairness and Accuracy}
Since fairness may hurt accuracy because it diverts a system's learning objective from accuracy only to both accuracy and fairness, we will assess each proposed model based on this trade-off. Figure~\ref{fig:compas_lambda} and Figure~\ref{fig:flc_lambda} show the fairness and accuracy trade-off plots for the FCPH models on the development set of the \emph{COMPAS} and \emph{FLC} datasets, respectively. The \emph{C-index} is selected as the accuracy-based performance measure in this experiment. The impact of the tuning parameter $\lambda$ on the C-index and corresponding fairness measures for the proposed FCPH models are demonstrated in these figures. Larger $\lambda$ values allow us to learn more fair, but less accurate FCPH models, while smaller $\lambda$ values have the opposite impact on the FCPH models.  

The tuning parameter $\lambda$ needs to be chosen as a trade-off between the \emph{C-index} and fairness. We chose $\lambda$ for all FCPH models via grid search on the development set based on a pre-defined rule: \emph{select the $\lambda$ that provides the fairest (under the corresponding fairness metric, e.g. $F_i$, $F_g$, and $F_\epsilon$ for \emph{individual}, \emph{group}, and \emph{intersectional FCPH} models, respectively ) Cox model on the development set, allowing up to $5\%$ degradation in \emph{C-index} from the typical CPH model}. In Figure~\ref{fig:compas_lambda} and Figure~\ref{fig:flc_lambda}, the \emph{red} square represents the typical CPH model without fairness penalty and the \emph{black} circles (correspond to different $\lambda$ values) above the \emph{dotted} line are FCPH models that degrade the \emph{C-index} but not over $5\%$. Finally, \emph{blue} squares indicate the selected fairest FCPH model for a specific $\lambda$ value that complies with the pre-defined rule.    

\subsection{Performance for FCPH Models} 
We evaluated the performance for FCPH models on the test data in terms of accuracy-based performance measures and proposed fairness measures, and compare our fair models with the typical CPH model. The goals of our experiments were to demonstrate the practicality of our FCPH models. 
In Table~\ref{fig:compas} and Table~\ref{fig:flc}, we show detailed results for \emph{COMPAS} and \emph{FLC} datasets, respectively. All FCPH models outperform typical CPH models in terms of all three fairness measures for both datasets. Note that the $\lambda$ values for FCPH models in Table~\ref{fig:compas} and Table~\ref{fig:flc} represent the \emph{blue} square from Figure~\ref{fig:compas_lambda} and Figure~\ref{fig:flc_lambda}, respectively.  In the \emph{COMPAS} dataset, the \emph{Individual FCPH} model was the most fair model in terms of $F_i$ measures, while \emph{intersectional FCPH} model was the most fair model in terms of $F_g$ and $F_\epsilon$ measures. The \emph{Individual FCPH} model shows superior performance on the \emph{FLC} dataset that outperforms all models in terms of all three fairness measures. The \emph{Group FCPH} model gives middling performance, and cannot win over the \emph{Individual} and \emph{intersectional FCPH} models in terms of fairness. This is presumably due to the fact that ensuring fairness for individuals or intersectional subgroups imposes a harder constraint to the objective function that automatically ensures fairness for groups. As expected, typical CPH performed best in terms of \emph{C-index} and \emph{Log Partial Likelihood}. However, surprisingly, we found that FCPH models also outperform typical CPH models in accuracy-based performance measures such as \emph{Brier Score} and \emph{AUC} (see \emph{group} and \emph{intersectional FCPH} models in Table~\ref{fig:compas}, and \emph{individual FCPH} model in Table~\ref{fig:flc}).     

\subsection{Do Fair Models Reduce Overfitting?} 
The improved performance of FCPH models over typical CPH models is counter-intuitive.   
We further study this result in this section by comparing the generalization of the models. Table~\ref{fig:compas_train} and Table~\ref{fig:flc_train} compare the accuracy-based predictive performances for FCPH models with typical CPH models on the train and test set of both datasets.

The typical CPH model was the best model for both datasets in all predictive measures on the training set, but FCPH models performed better than typical CPH models on the test set  for both datasets in most of the cases. In the \emph{COMPAS} dataset, \emph{group} and  \emph{intersectional FCPH} models were the best models on the held-out data in terms of \emph{AUC} and \emph{Brier Score}, respectively. Similarly, \emph{individual FCPH} models showed the best \emph{Brier Score} and \emph{AUC} measures on the held-out \emph{FLC} data. We also found that FCPH models decrease the corresponding gap between accuracy-based predictive measures on train and test data due to the regularization behavior of the fairness constraints. Thus, FCPH models reduce overfitting of the typical Cox model to some extent.   

\section{Discussion and Future Work}
In this work, we investigated fairness for survival models and developed methods to ensure fair risk scores. Balance between accuracy and fairness is an important decision to make prior to learning fair models, and this depends on the stakeholders. To validate our proposed fair models, we performed all the experiments on two public proxy datasets. 

In future, we plan to apply our proposed methods on the data from healthcare programs such as Medicaid to ensure fair risk prediction in  ranking the waitlist for individuals to receive care. We further plan to study the impact of an intervention to the prioritization process on each individual which determines the waiting time to receive home and community-based healthcare services. The successful application and deployment of our methods to the healthcare program is the eventual goal of this research. 

\section{Conclusion}
We developed three fairness definitions for survival models and corresponding learning algorithms to ensure equitable allocation of healthcare resources. In extensive experiments on publicly available datasets, we demonstrated that our methods are practical and effective. The proposed methods for fair prioritization of healthcare have the potential to prevent avoidable institutionalization of elderly and disabled individuals, thereby improving quality of life and saving taxpayer dollars, while ensuring fair and equitable allocation.

\section{ Acknowledgments}
This material is based
upon work supported by the National Science Foundation under
Grant No.'s IIS1927486; IIS1850023. Any opinions, findings, and
conclusions or recommendations expressed in this material are
those of the author(s) and do not necessarily reflect the views of
the National Science Foundation. This work was performed under the following financial assistance award: 60NANB18D227 from U.S. Department of Commerce, National Institute of Standards and Technology.
\bibliographystyle{aaai}
\bibliography{references}

\end{document}